\documentclass[11pt]{article} 
\usepackage{rldmsubmit,palatino}
\usepackage{graphicx}
\usepackage[comma,numbers]{natbib}
\usepackage{amsmath}
\usepackage{mathpazo}

\usepackage[colorlinks=true, allcolors=blue]{hyperref}

\title{Is there \textit{Value} in Reinforcement Learning?} 

\author{
Lior Fox\\
Gatsby Computational Neuroscience Unit\\
University College London\\
\texttt{lior.fox@ucl.ac.uk} \\
\And
Yonatan Loewenstein\\
Edmond and Lilly Safra Center for Brain Sciences\\
Hebrew University, Jerusalem\\
\texttt{yonatan@huji.ac.il} 
}

\newcommand{\E}{\operatornamewithlimits{\mathbb{E}}}

\begin{document}

\maketitle

\begin{abstract} 

Action-values play a central role in popular Reinforcement Learing (RL) models of behavior. Yet, the idea that action-values are explicitly represented has been extensively debated. Critics had therefore repeatedly suggested that policy-gradient (PG) models should be favored over value-based (VB) ones, as a potential solution for this dilemma. Here we argue that this solution is unsatisfying. This is because PG methods are not, in fact, ``Value-free'' -- while they do not rely on an explicit representation of Value for \textit{acting} (stimulus-response mapping), they do require it for \textit{learning}. Hence, switching to PG models is, \textit{per se}, insufficient for eliminating Value from models of behavior. More broadly, the requirement for a representation of Value stems from the underlying assumptions regarding the optimization objective posed by the standard RL framework, not from the particular algorithm chosen to solve it. 
Previous studies mostly took these standard RL assumptions for granted, as part of their conceptualization or problem modeling, while debating the different methods used to optimize it (i.e., PG or VB). We propose that, instead, the focus of the debate should shift to critically evaluating the underlying modeling assumptions. Such evaluation is particularly important from an experimental perspective. Indeed, the very notion of Value must be reconsidered when standard assumptions (e.g., risk neutrality, full-observability, Markovian environment, exponential discounting) are relaxed, as is likely in natural settings.
Finally, we use the Value debate as a case study to argue in favor of a more nuanced, algorithmic rather than statistical, view of what constitutes ``a model'' in cognitive sciences. Our analysis suggests that besides ``parametric'' statistical complexity, additional aspects such as computational complexity must also be taken into account when evaluating model complexity.

\end{abstract}


\acknowledgements{We thank Lotem Elber-Dorozko and Ohad Dan for discussions, and for critically evaluating an earlier version of this text.\\This work was supported by the Gatsby Charitable Foundation.}

\startmain 

\section{Reinforcement Learning models}

The Reinforcement Learning (RL) framework has been widely used to model the behavior of animals and humans in decision-making tasks (for reviews, see \cite{dayan2008rlreview,niv2009reinforcement, mongillo2014misbehavior, fox2020exploration}). The formalism conveniently fits into experimental setups: it consists of an agent-environment interaction loop, characterized by states (of the environment), actions (taken by the agent, by observing the current state and choosing an action from its policy), and rewards (associated with state-action pairs). Besides the descriptive element, RL models typically come with a normative assumption regarding the agent's goal -- maximizing expected cumulative (and often temporally discounted) reward.

So-called ``model-free'' RL algorithms provide the powerful guarantee that even if the rules underlying state-transitions and rewards are unknown to the agent, it can still learn to act optimally, even without trying to explicitly learn aforementioned rules. These algorithms largely fall into two families, value-based (VB) and policy-gradient (PG). VB methods are based on the insight that the notion of optimality defined above admits a \textit{Dynamic Programming} solution \citep{bellman1957}, through the well-known Bellman optimality equations. Solving the optimality equations explicitly would require an access to the state-transition and reward rules (the ``model''). Instead, VB methods solve these implicitly -- by online approximating the dynamic programming solution from observations. This typically relies on temporal-difference (TD) style learning driven by reward prediction error. A key structure in these algorithms is, as the name suggests, the policy's action-values. These measure, for any given state-action pair, the expected cumulative reward achieved by starting at this state-action, and then following the policy for all future choices. Therefore, by construction, a policy for which the action-values are maximal will solve the optimization problem.

PG methods, in their basic form, do not explicitly make use of the dynamic programming structure. Instead, they solve the optimization problem by calculating, from observations, a stochastic estimate of the gradient of the objective function with respect to the set of parameters defining a policy, and update the parameters in the direction of this (estimated) gradient \citep{williams1992simple, sutton1999PG}.

\section{The arguments against Value}
Despite (and, perhaps, also thanks to) their predominance, the extent to which RL algorithms are good models of natural behavior remains debated \citep{dan2019choice,rosenberg2021maze, dan2025behavior}. Within the larger debate, the notion of an explicit representation of Value has been particularly contested on several different levels. Criticism has been made of both the assumption that biological behavior is primarily guided by top-down optimality considerations \citep{plonsky2017catie,suri2020value}, as well as the assumption of a ``common currency'' that is used for (numerical) evaluation of different actions \citep{hayden2021case}. Moreover, it has been shown that the evidence for Value representation in the brain is weaker than previously realized \citep{li2011pg,o2014problem,elber2018value}. Finally, the stimulus-response mapping in VB models is ``indirect'', relying on the hypothesized (and latent) Value variables, internal to the agent \citep{bennet2021valuefree}. 

Following these criticisms, several authors have argued that PG models should be favored as the go-to modeling approach of biological behavior. The basis for these arguments is that PG methods could account for most experimental observations attributed traditionally to VB methods, while being conceptually simpler, requiring less assumptions, and (so goes the argument) eliminate the questionable Value construct from the models \citep{mongillo2014misbehavior,hayden2021case, bennet2021valuefree}.

At a first glance, the suggestion is appealing. Indeed, PG methods are more ``direct'': they directly update the variables defining the optimization problem (the policy). They might also look appealing from the perspective that it is often (much) easier finding out \textit{which} action is better, than by precisely \textit{how much} it is better (i.e., evaluating numerically the advantage that one action has over another) \citep{simsek2016tetris, laidlaw2023bridging}. In the next section, however, we will argue that switching to a PG model doesn't answer, \textit{per se}, the criticisms outlined before. This is simply because the switch to a PG model does \textit{not} eliminate the requirement for a representation of Value. 

Before moving on to our main argument, it should be made clear that we are not trying to argue that the criticisms made against Value are somehow irrelevant. On the contrary, we believe many of these important, and call for a critical re-assessments of RL modeling in neuroscience, as we will detail later.

\section{Policy Gradient Methods are not ``Value-free''}

In basic forms of PG, the agent does not maintain a persistent representation of values that is used for \textit{acting}; rather, it directly parameterizes a policy which maps states to actions. Nevertheless, a representation of Value is required for \textit{learning}, in order to guide the updates for this policy. Intuitively, PG can be thought of as rolling-out actions according to the current policy, then updating the policy such that the probability of successful actions is increased the next time around (and conversely for unsuccessful actions). Crucially, how ``successful'' a particular action was is measured by its Value. Formally, this is evident in the well-known PG theorem \citep{sutton1999PG}:
\begin{equation}\label{eq:pg_theorem}
	\nabla_\theta J\left(\pi_\theta\right) = \E_{s,a}\left[Q^{\pi_\theta}\left(s,a\right)\nabla_\theta\log\pi_\theta\left(s,a\right)\right]
\end{equation}
Where $\pi_\theta$ is a policy parameterized by $\theta$, $J(\pi_\theta) = \E_{\pi_\theta}\left[\sum_{t=0}^\infty \gamma^t r(s_t,a_t)\right]$ is the objective function, $r$ is the reward function, $Q^{\pi_\theta}$ is the (state-action) value function of the policy, and $\gamma$ is the discount factor.

Different PG algorithms differ in how they estimate the Values needed for the update rule. For example, in \textsc{reinforce} \citep{williams1992simple}, a simple sampling-based estimate (observed cumulative rewards) is used, resulting in the following learning-rule: 
\begin{equation}\label{eq:reinforce}
	\nabla_\theta J(\pi_\theta) \approx \sum_t \nabla_\theta\log\pi_\theta(a_t|s_t)R_t
\end{equation} 
where $R_t$ is the \textit{empirical} return at time $t$, collected in an observed trajectory, namely $	R_t=\sum_{\tau=0}^\infty \gamma^\tau r(s_{t+\tau},a_{t+\tau})$. Note that the quantity $R_t$ is, by definition, a sampling-based estimate of $Q^{\pi_\theta}(s_t,a_t)$.

Crucially, $R_t$ cannot be replaced by $r(s_t,a_t)$ in the learning rule: such modification will, in general, lead to learning a sub-optimal policy \citep{loewenstein2006matching}. One special case in which this replacement \textit{is} valid is in multi-armed bandit problems. Because bandit problems are extremely common in behavioral experiments, it is not uncommon for authors to analyze and demonstrate different approaches in a bandit. But this might have contributed to a misleading interpretation that in PG methods, the immediate reward is the direct learning signal or regulator: the only reason this holds in bandit problems, is that in those problems (expected) reward and value are one and the same. One (tangential) lesson is that in order to fully appreciate the implications and predictions of RL models in experimental works, it is important to analyze them in MDPs that are richer than bandit problems alone \citep{fox2023comp}.

In this sense, PG methods are not ``Value-free'': the agent must still represent action-values as part of its learning.\footnote{In some PG variants, value is not \textit{directly} represented, and instead eligibility traces are used to track visits of state-actions throughout the trajectory \citep[e.g.][]{baxter2001infinite}. This is conceptually similar to way that a value-function can be computed from the successor representation (SR) \citep{dayan1993improving}. Note that in both cases, value can be read-out directly by some mathematical transformation. These PG variants (unlike the SR) have not been popular in the behavioral RL field, hence we focus our discussion on the more common, \textsc{reinforce}-like, variants.} The main technical difference is that these action-value representations are ``volatile'' -- they are not kept from one episode to the next, and are not being used to update or maintain some persistent representation of value that can be queried by the policy at acting time. 

It is instructive to consider VB methods following the same considerations. While in these models there is no independent representation of the policy, they are not ``policy free'' -- they simply derive a behavioral policy from the estimated values. That is, their policy component is ``volatile'', rather than an updated, persistent representation. The complementary aspect is that while behavior (the mapping of stimulus to response; states to actions) in VB models is indirect, compared to PG models, they can potentially have direct learning (the mapping of stimulus to change in parameters), while learning is more indirect in PG models. This is because in VB models, learning can truly be fully online -- each observation from the environment (a transition from state-action to state-action along with the observed reward) instantaneously drives an update of the (estimated) value function, and the agent does not have to maintain a trace of entire trajectories in memory. 

Viewed this way, PG and VB methods can both be understood as an instance of the general RL technique that Sutton and Barto termed Generalized Policy Iteration (GPI) \cite{sutton2018rl2nd}. GPI consists of two steps that are performed iteratively. One step is to measure, estimate, or compute the performance of the current policy. The second is to modify the policy in such a way as to (perhaps slightly) increase this performance. These steps are often called policy \textit{Evaluation} and policy \textit{Improvement}. 
The notions of policy evaluation and improvement are tightly related to those of \textit{actor} and \textit{critic} ``modules'' in an agent architecture. 
Today, the term ``Actor-Critic'' had become associated with a particular style of RL algorithms, explicitly combining PG and Value-based methods. 
Here we use these terms more broadly, to denote any component of the agent or architecture involved in improvement and evaluation (in fact, the idea of such double architecture predates both VB and PG methods, e.g. \cite{barto1983adaptive}; for historical review see \citep{baxter2001infinite}). With this interpretation, ``pure'' VB and PG methods can both be understood as forms of Actor-Critic, which differ in the way they implement the abstract idea of GPI. In PG the ``actor'' is explicitly modeled parametrically, while the ``critic'' is a simple, non-parametric and volatile component. In VB methods the ``critic'' is explicitly modeled parametrically, while the ``actor'' is a simple, volatile, derived mapping of the critic.

\section{Are theories of Value necessary?}
Within the context of the standard RL framework (maximizing expected cumulative discounted reward in a Markovian environment), the fact that Value appears in both PG and VB methods should, perhaps, come as no surprise. After all, both methods attempt to optimize the same objective, and this objective is \textit{defined} in terms of Value. Thus, if one aims at eliminating, or at least reconsidering, this variable from models of operant learning, it is crucial to start at the foundations: the posited optimization problem, rather than at the details of the particular optimization methods. Moreover, from a theoretical perspective, it is becoming clearer that PG and VB methods are not as distinct as traditionally appreciated. Not only do they converge (in general) at the same solution of the optimal policy, but also, in some cases, aspect of their transient learning dynamics can be mapped to each other \citep{kakade2001npg, nachum2017bridging}.

More fundamentally, the existence of a Value function relies on the Bellman Equations, which in turn depend on a set of non-trivial assumptions about the problem. These are often taken for granted by adopting standard RL as a conceptual framework for studying behavior. Yet, in the learning and behavior of biological agents, all of these assumptions (for example exponential discounting and risk neutrality) have been challenged before, and their violations preclude the existence of Value in the first place. If these assumptions are relaxed, significant modifications must be made in order for any algorithm to solve the modified optimization problem(s). Recent studies in RL theory have demonstrated that basic insights from both methods can be modified in this way. For example, PG methods can still work as long the agent is able to evaluate how ``successful'' an entire trajectory is (even if this cannot be decomposed into a sum of evaluations of individual actions) \citep{zhang2020variationalPG}, or in partially observable scenarios, where alternative notions of optimality are considered \citep{loewenstein2006matching,loewenstein2009nash}. Similarly, VB methods can also be adapted to handle non-standard cases, provided a more general notion of Value (and even of the basic underlying ``reward'' generating the Value function) is being used \citep{zahavi2021convexrl}.

We therefore conclude that it is instructive to shift our attention to questions about the nature and properties of Value as implied by the behavior (or neural activity) of biological agents, rather than asking whether or not Value in its classical sense is being represented. This shift might help resolve some, though not all, of the criticisms against Value representation in the brain. Even if Value is maintained in the models, going beyond the superficial PG-or-VB debate could help emphasize a new set of questions approachable within the RL framework, that had previously received less attention, but have potentially important implications for the psychological and neural basis of operant learning. These questions might include, for example, temporal-difference (``bootstrapping'') versus monte-carlo methods, 1-step backups versus trajectory-based updates, and on-policy versus off-policy learning.

\section{What's in a model?}

We conclude by taking another look at the question of model simplicity. It is a tenet of scientific thinking that when comparing competing explanations for a phenomenon, simpler explanations should be preferred over complex ones (provided both are able to explain the phenomenon). As discussed earlier, this was one of the arguments proposed for favoring PG over VB methods \citep{bennet2021valuefree}.

And yet, there is no universal principle for determining an explanation's (or model's) complexity. A widely accepted practice in contemporary brain and behavioral sciences is to adopt a \textit{statistical} notion of complexity, relying on the number of free parameters that have to be estimated from data. 
These criteria on their own ignore, however, other aspects of complexity that are just as relevant in modeling cognitive phenomena, such as \textit{computational} complexity.

Consider again a PG and a VB algorithm for modeling behavior. From a statistical perspective, the VB model might require additional parameters (responsible for the mapping between Value, which is latent, and actions, which are observed) compared to the PG model. However, from a space-complexity perspective, VB model is simpler: the agent can learn in a fully online way, without storing entire trajectories in memory, as is required by the PG model. VB methods do not gain this simplicity for free: as previously discussed, these methods make more explicit assumptions about the problem, so they can offload parts of the complexity to these assumptions (specifically, the assumptions that enable a dynamic programming solution). These extra assumptions can be viewed as yet \textit{another} axis of model complexity.

There might not be an ``off-the-shelf'' method to account for such multi-axis complexity, but acknowledging it is nonetheless important. Computational models such as RL algorithms are often interpreted as \textit{mechanistic}, at least on the level of the cognitive architecture (and often also at the level of neural implementation). If such interpretations are to be taken seriously, then all algorithmic components of the model, including ``non-parametric'' ones (e.g., the sampling-based estimation of Value in \textsc{reinforce}) must be factored-in when evaluating model complexity. A PG \textit{agent} is not just the neural network (or lookup table) policy, and a VB agent is not just the value neural network (or table): they are entire \textit{algorithms} for updating, manipulating, and using these networks, and it is these algorithms which serve, effectively, as our model for the biological organism in the experiment. We believe that this more holistic interpretation of ``a model'' is needed in other domains where machine-learning algorithms are serving as quantitative models in cognitive sciences.

\bibliography{sources}

\end{document}